%% file: main.tex
\documentclass[letterpaper, 10 pt, conference]{ieeeconf}
\IEEEoverridecommandlockouts
\usepackage{cite, balance}
\usepackage{amsmath,amssymb,amsfonts}

\usepackage{amsthm}
\usepackage{algorithmic}
\usepackage{mathtools}
\usepackage[font=small,skip=0pt]{caption}
\DeclareMathOperator*{\argmax}{arg\,max}

\usepackage{babel}
\usepackage{xurl}
\usepackage{multirow}
 
\input{preamble}

\theoremstyle{definition} 
\newtheoremstyle{italdefinition} 
{} 
{} 
{} 
{} 
{\itshape} 
{:} 
{3pt} 
{} 
\theoremstyle{italdefinition}
\newtheorem{definition}{Definition}

\newcommand{\yifei}[1]{\textcolor{black}{#1}}

\usepackage{caption}
\usepackage{ifthen}
\newboolean{includeFigures}
\setboolean{includeFigures}{true}  
\newboolean{includeAppendix}
\setboolean{includeAppendix}{false}
\usepackage{subcaption}
\usepackage{float}
\usepackage{booktabs}
\usepackage{array}

\makeatletter
\renewcommand\paragraph{\@startsection{paragraph}{4}{\z@}%
                                    {0.3ex \@plus1ex \@minus.2ex}%
                                    {-1em}%
                                    {\normalfont\normalsize\bfseries}}
\makeatother

\usepackage[utf8]{inputenc}
\usepackage{pgfplots}
\DeclareUnicodeCharacter{2212}{−}
\usepgfplotslibrary{groupplots,dateplot}
\usetikzlibrary{patterns,shapes.arrows}
\pgfplotsset{compat=newest}
\pgfplotsset{every tick label/.append style={font=\tiny}}
\usetikzlibrary{shapes.geometric}
\usetikzlibrary{calc}

\usepackage[linesnumbered,ruled,vlined]{algorithm2e}
\usepackage{amssymb}

\usepackage{hyperref}
\hypersetup{
    colorlinks=true,
    linkcolor=blue,
    filecolor=magenta,      
    urlcolor=blue,
    pdftitle={CageCoOpt: Enhancing Manipulation Robustness through Caging-Guided Morphology and Policy Co-Optimization},
    pdfpagemode=FullScreen,
    }
\usepackage{graphicx}
\usepackage{textcomp}
\usepackage{xcolor}
\def\BibTeX{{\rm B\kern-.05em{\sc i\kern-.025em b}\kern-.08em
    T\kern-.1667em\lower.7ex\hbox{E}\kern-.125emX}}

\title{\LARGE \bf
CageCoOpt: Enhancing Manipulation Robustness through Caging-Guided Morphology and Policy Co-Optimization 
}

\author{Yifei Dong$^{1,*}$, Shaohang Han$^{1,*}$, Xianyi Cheng$^{2}$, Werner Friedl$^{3}$, Rafael I. Cabral Muchacho$^{1}$, \\Máximo A. Roa$^{3}$, Jana Tumova$^{1}$ and Florian T. Pokorny$^{1}$ 
\thanks{$^{*}$ These authors contributed equally.}
\thanks{$^{1}$ The authors are with the division of Robotics, Perception and Learning, KTH Royal Institute of Technology, 10044 Stockholm, Sweden. 
$^{2}$ The author is with the Department of Mechanical Engineering and Material Science, Duke University, Durham, NC 27708, USA.
$^{3}$ The authors are with the Institute of Robotics and Mechatronics, German Aerospace Center (DLR), 82234 Wessling, Germany.
Funded by the European Commission under the Horizon Europe Framework Programme project SoftEnable, grant number 101070600. Contact: {\tt\small \{yifeid, shaohang\}@kth.se}.
}
}

\begin{document}

\maketitle
\thispagestyle{empty}
\pagestyle{empty}

\begin{abstract}
\yifei{
Uncertainties in contact dynamics and object geometry remain significant barriers to robust robotic manipulation. Caging mitigates these uncertainties by constraining an object's mobility without requiring precise contact modeling. However, existing caging research has largely treated morphology and policy optimization as separate problems, overlooking their inherent synergy. In this paper, we introduce CageCoOpt, a hierarchical framework that jointly optimizes manipulator morphology and control policy for robust manipulation. The framework employs reinforcement learning for policy optimization at the lower level and multi-task Bayesian optimization for morphology optimization at the upper level. A robustness metric in caging, Minimum Escape Energy, is incorporated into the objectives of both levels to promote caging configurations and enhance manipulation robustness. The evaluation results through four manipulation tasks demonstrate that co-optimizing morphology and policy improves success rates under uncertainties, establishing caging-guided co-optimization as a viable approach for robust manipulation.
}
\end{abstract}



\section{Introduction}

\yifei{Robust non-prehensile manipulation under real-world uncertainties has been a long-standing challenge in robotics. These uncertainties arise from various sources, including imperfect sensors and actuators, varying object geometrical and physical properties, complex contact dynamics, etc~\cite{bhatt2022surprisingly}. In model-based manipulation planning, discontinuous contact interactions make it difficult to maintain robustness, as small deviations in contact conditions can significantly alter system behavior.
}

\yifei{Caging offers a promising approach to mitigating these uncertainties by constraining an object's mobility without requiring precise contact modeling. A manipulator can act as a set of geometric constraints, forming a ``cage'' that constrains the movement of the object~\cite{kuperberg1990problems}. Unlike force- or form-closure grasps, caging allows for bounded mobility while ensuring that the object remains under control, thereby enhancing robustness to geometric and positional uncertainties. Existing research on caging has primarily focused on theoretical analysis~\cite{rodriguez2012caging, shirizly2024selection, varava2017herding}, robotic hand design~\cite{bircher2021complex, beddow2021caging, bircher2019design} and manipulation planning~\cite{wang2024caging, mahler2018synthesis}. However, the synergy between caging-based morphology and policy optimization remains an open problem, while they are strongly connected. The manipulator’s morphology directly influences feasible control policy, while the policy must adapt to the constraints and affordances imposed by its morphology.
}

\yifei{In light of this, we introduce CageCoOpt, a hierarchical framework that co-optimizes the manipulator morphology and policy for caging-based robust manipulation (Fig.~\ref{fig:teaser}). The framework employs reinforcement learning for lower-level policy optimization and multi-task Bayesian optimization for upper-level morphology optimization. To account for uncertainties, we explicitly incorporate object shape variation and environmental disturbances into both the learning and evaluation processes.
}

\ifthenelse{\boolean{includeFigures}}{
\begin{figure}[t]
    \centering
    \includegraphics[width=1\linewidth]{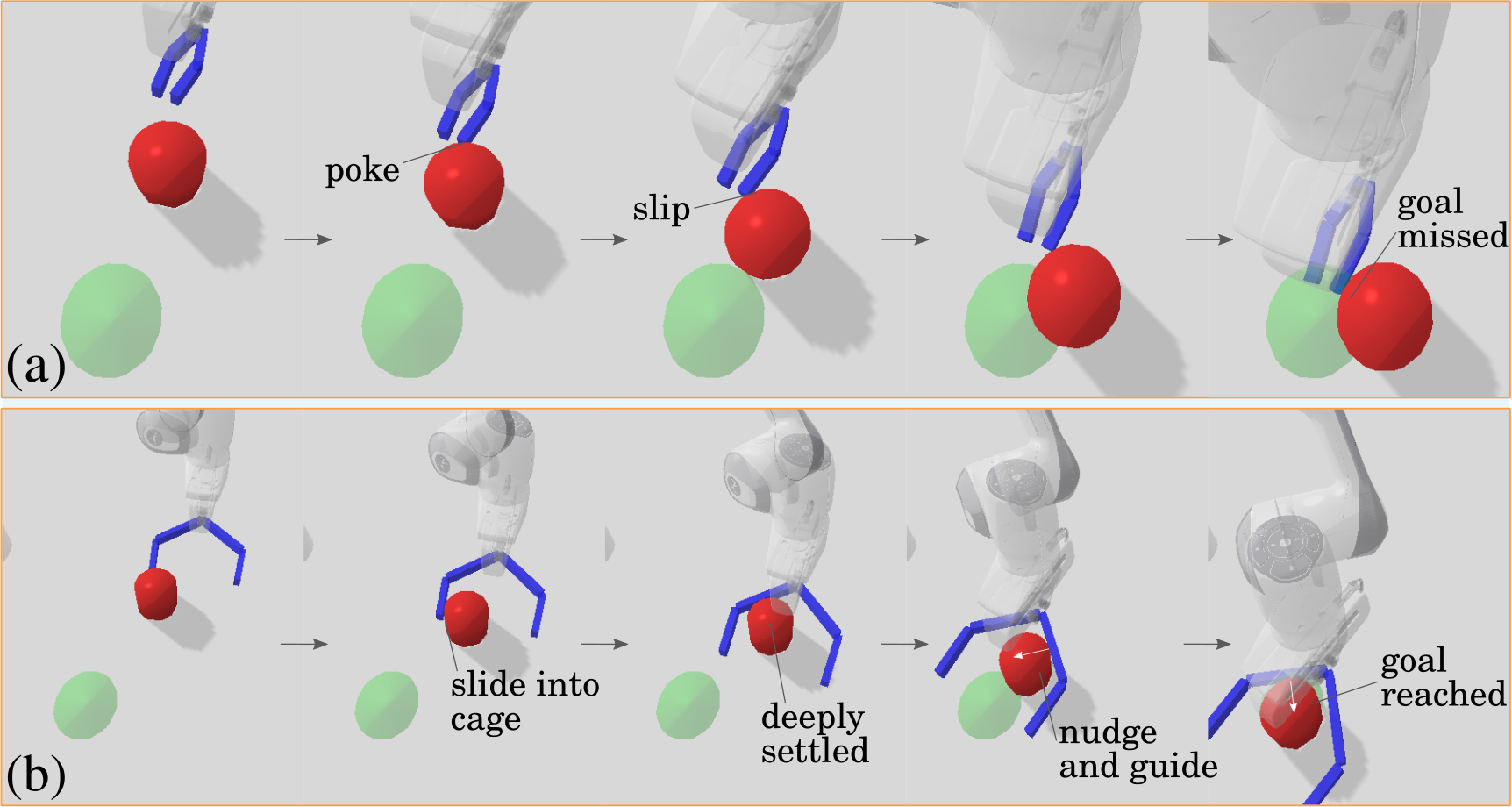}
    \caption{
        We introduce \textit{CageCoOpt}, a framework that jointly optimizes manipulator \textit{morphology} and control \textit{policy} for \textit{caging}-based manipulation. 
        The goal is to push an object (red) to the target region (green).
        (a) A non-optimized manipulator (blue) struggles with pushing the object reliably. Fingertip poking often fails due to contact slippage, as precise control is required to handle object shape variations and unmodeled dynamics.  
        (b) Through CageCoOpt, the manipulator evolves into a partially ``cage''-like morphology, while the policy learns to nudge the object using its inner surface. The manipulator maintains partially caging the object and guides it to the goal without reliance on precise contacts. This co-optimized system adapts to diverse object shapes and external disturbances, ensuring robust manipulation.  
    }
    \label{fig:teaser}
\end{figure}
}{}

\yifei{\textbf{Contributions}: We integrate Minimum Escape Energy (MEE), a robustness metric from energy-bounded caging~\cite{mahler2018synthesis, dong2024quasi}, into the optimization objectives of the two levels. By embedding MEE in the learning process, our method encourages caging configurations. Specifically, the manipulator evolves into a morphology that cages the object while the policy adapts to maintain this cage. To the best of our knowledge, this is the first instance in which the morphology and policy of manipulators are jointly optimized to facilitate caging-based manipulation. Furthermore, we evaluate CageCoOpt in both simulation and real-world experiments across four non-prehensile manipulation tasks, such as pushing and scooping. The results demonstrate that integrating MEE into morphology and policy optimization consistently improves success rates under uncertainties. The manipulator with jointly optimized morphology and policy outperforms its unoptimized counterparts in handling real-world uncertainties.
}

\section{Related Work}
\subsection{Robust Manipulation and Caging}
\label{sec:2-robmanip}
\yifei{Early work on manipulation robustness explores various grasp quality metrics, such as grasp wrench space analysis~\cite{ferrari1992planning, pollard1996synthesizing}, with a primary focus on prehensile manipulation tasks~\cite{roa2015grasp}. Data-driven approaches also leverage analytic metrics for planning robust grasps~\cite{saxena2008learning, mahler2017dex, mahler2018dex}. However, robustness in non-prehensile manipulation remains underexplored. In general, uncertainties in manipulation arise from various sources, including objects, the environment, robots, or the interactions between them~\cite{wang2020feature, bhatt2022surprisingly}. Of particular interest are two challenging types of uncertainty: object shape variation and unmodeled dynamics. Object geometric uncertainty stems from the limited sensing capabilities of robots in unknown environments~\cite{bohg2013data, daniels2023grasping}, while unmodeled dynamics can result from contact dynamics between objects and obstacles, or from uncertainties in the physical properties of objects~\cite{andrychowicz2020learning, jankowski2024planning}.}

\yifei{Caging~\cite{kuperberg1990problems, rimon1999caging, rodriguez2012caging} and energy-bounded caging~\cite{mahler2018synthesis, dong2024quasi, dong2024characterizing, shirizly2024selection}, a variant of caging that incorporates energy constraints on the object, provide valuable insights into addressing these challenges. Caging involves preventing object escape through geometric constraints without relying on force or form closure, allowing for the tolerance of geometric or dynamic uncertainties.
In this work, we take initial steps toward integrating caging-based analytical metrics into the co-optimization of morphology and policy. This approach offers a paradigm for applying analytical manipulation metrics in a data-driven manner to enhance non-prehensile manipulation performance, complementing state-of-the-art research on caging for independent robot design~\cite{bircher2019design, bircher2021complex, beddow2021caging} or manipulation planning~\cite{varava2017herding, mahler2018synthesis, wang2024caging}.}

\subsection{Simultaneous Morphology and Policy Optimization}
\label{sec:2-codesign}
\yifei{Simultaneous morphology and policy optimization, known as co-design or co-adaptation of robots, reduces costly manual hardware design by integrating policy into the adaptation process. Co-design is a bi-level optimization problem~\cite{zhang2024introduction}, involving upper-level design and lower-level control optimization. This approach is computationally challenging, as each morphological design requires a unique policy, exacerbating the complexity with increasing morphology space dimensionality. It has been applied to soft robots~\cite{wang2024diffusebot, bhatia2021evolution}, manipulation~\cite{kim2021mo, liu2023learning, schneider2024task}, locomotion~\cite{rajani2023co, yuan2022transform2act, jackson2021orchid}, and object adaptation~\cite{liu2024paperbot, guo2024learning}. Prior work falls into hierarchical and integrated approaches. Hierarchical methods use Bayesian optimization~\cite{kim2021mo, pan2021emergent, rajani2023co, schneider2024task}, evolutionary algorithms~\cite{bhatia2021evolution}, or generative methods~\cite{wang2024diffusebot, xu2024dynamics} for morphology optimization, with reinforcement learning~\cite{bhatia2021evolution, schneider2024task} or imitation learning~\cite{rajani2023co} for policy optimization. Some simplify control as prescribed motions~\cite{kim2021mo, xu2024dynamics, liu2024paperbot}. Integrated approaches frame co-design as a joint Markov Decision Process (MDP)~\cite{liu2023learning, yuan2022transform2act, jackson2021orchid, guo2024learning}, simplifying the pipeline but limiting flexibility by restricting optimization to a single objective. Notably, existing co-design methods do not systematically address robust manipulation under uncertainties. To tackle this, we propose a hierarchical framework with decoupled subproblems: a morphology-conditioned control policy is first learned through reinforcement learning, followed by efficient morphology optimization using multi-task Bayesian optimization~\cite{swersky2013multi}. This allows for customized objectives, integrating caging-based robustness metrics independently at each level.
}

\section{Preliminary}
\label{sec:2}
We use the vertical bar \( | \) in expressions like \( f(a | b) \) to indicate parameter dependency, meaning \( f \) is a function of \( a \) with \( b \) as a fixed parameter influencing its behavior.

We begin by revisiting the concept of caging. $\mathcal{S}_{\text{obj}}$ represents a rigid object's configuration space (C-space). The free C-space, $\mathcal{S}_{\text{free}}$, consists of configurations where the object does not intersect any obstacles or the manipulator.

\begin{definition}
    An object at configuration $s_{\text{obj}} \in \mathcal{S}_{\text{free}}$ is \textit{caged} if it resides in a bounded path component of $\mathcal{S}_{\text{free}}$.
    \label{def1}
\end{definition}

A caged object is restricted in its movement and unable to travel freely beyond a certain range from its initial configuration. Caging has been generalized to partial caging, which permits escape through narrow free-space passages~\cite{makapunyo2012measurement}. This extends to energy-bounded caging~\cite{mahler2018synthesis, dong2024quasi}, incorporating both geometric and energy constraints.

\begin{definition}
     In a quasi-static system under conservative forces, an object at an initial configuration $s_{\text{obj}}\in\mathcal{S}_{\text{free}}$ has potential energy $E(s_{\text{obj}})$. The \emph{Minimum Escape Energy} $\mathcal{E}(s_{\text{obj}})$ is the supremum of energy gain $e$ such that the path component $\mathcal{PC}_{s_{\text{obj}}} (\mathcal{S}_{e}(s_{\text{obj}}))$ of the sublevel set $\mathcal{S}_{e}(s_{\text{obj}})$ containing $s_{\text{obj}}$ remains bounded:
    \begin{align} 
    \mathcal{E}(s_{\text{obj}}) &= \sup\left\{e \ge 0 : \mathcal{PC}_{s_{\text{obj}}} (\mathcal{S}_{e}(s_{\text{obj}}))  \text{ is bounded} \right\}, \label{eq0}
    \end{align}
    where
    \begin{equation} \label{eq1}
    \mathcal{S}_{e}(s_{\text{obj}}) = \{s \in \mathcal{S}_{\text{free}} : E(s) \le E(s_{\text{obj}}) + e \}.
    \end{equation}
    The object is in an \emph{$\mathcal{E}(s_{\text{obj}})$-energy-bounded cage}~\cite{mahler2018synthesis}. $\mathcal{E}(s_{\text{obj}})$ is defined only if $\mathcal{PC}_{s_{\text{obj}}} (\mathcal{S}_{0}(s_{\text{obj}}))$ is bounded.
\label{def2}
\end{definition}

The supremum of energy gain $e$ satisfying the condition in Def.~\ref{def2}, $e^{\text{sup}}=\mathcal{E}(s_{\text{obj}})$, is noted in Fig.~\ref{fig:ebc}. It corresponds to a maximal bounded path component $\mathcal{PC}_{s_{\text{obj}}} (\mathcal{S}_{e^{\text{sup}}}(s_{\text{obj}}))$ (blue). The object at configuration $s_{\text{obj}}$ is bounded by both the $\mathbb{R}^2$-bowl (obstacle C-space $\mathcal{S}_{\text{obs}}$) and the energy level set $\mathcal{L}_{e^{\text{sup}}}(s_{\text{obj}})$, defined as:
\begin{equation}
  \mathcal{L}_{e}(s_{\text{obj}}) = \{ s \in \mathcal{S}_{\text{free}} : E(s) = E(s_{\text{obj}}) + e \}.  
\end{equation}
\noindent In other words, it is in an ``energy-bounded cage'' within the blue region. For any $e>e^{\text{sup}}$, the path component $\mathcal{PC}_{s_{\text{obj}}} (\mathcal{S}_{e}(s_{\text{obj}}))$ of the sublevel set $\mathcal{S}_{e}(s_{\text{obj}})$ containing $s_{\text{obj}}$ is unbounded.
Intuitively, if an object at configuration $s_{\text{obj}}$ has positive Minimum Escape Energy (MEE), $\mathcal{E}(s_{\text{obj}}) > 0$, it is confined within a region around a local minimum of the energy function $E(s): \mathcal{S}_{\text{free}} \rightarrow \mathbb{R}$. It cannot escape without an external energy input greater than $\mathcal{E}(s_{\text{obj}})$. 
Otherwise, if $\mathcal{E}(s_{\text{obj}}) = 0$, the object can readily escape from the region.


\yifei{
Efficient sampling-based approaches such as Batch
Informed Trees (BIT*) \cite{gammell2015batch} are used in practice to approximate a close upper bound of MEE $\mathcal{E}(s_{\text{obj}})$.
Specifically, a C-space goal region $\mathcal{S}_{\text{goal}}$ (pink, Fig.~\ref{fig:ebc}) sufficiently far away from the obstacles with relatively lower energy values is utilized. With the help of it, we search for geometrically feasible \textit{escape paths} $\alpha$ from $s_{\text{obj}}$ to the goal region $\mathcal{S}_{\text{goal}}$ with the lowest energy cost. The introduction of sampling-based approaches for finding the most energy-efficient escape paths also makes it possible to extend the assumed conservative forces, such as gravity, to path-dependent non-conservative forces, such as friction~\cite{dong2024characterizing}.
Please refer to \cite{dong2024quasi, dong2024characterizing} for details. We next describe how we integrate MEE $\mathcal{E}(s_{\text{obj}})$ into both the two levels of our hierarchical co-optimization framework.
}

\ifthenelse{\boolean{includeFigures}}{
\begin{figure}[t]
    \centering
    \includegraphics[width=0.55\linewidth]{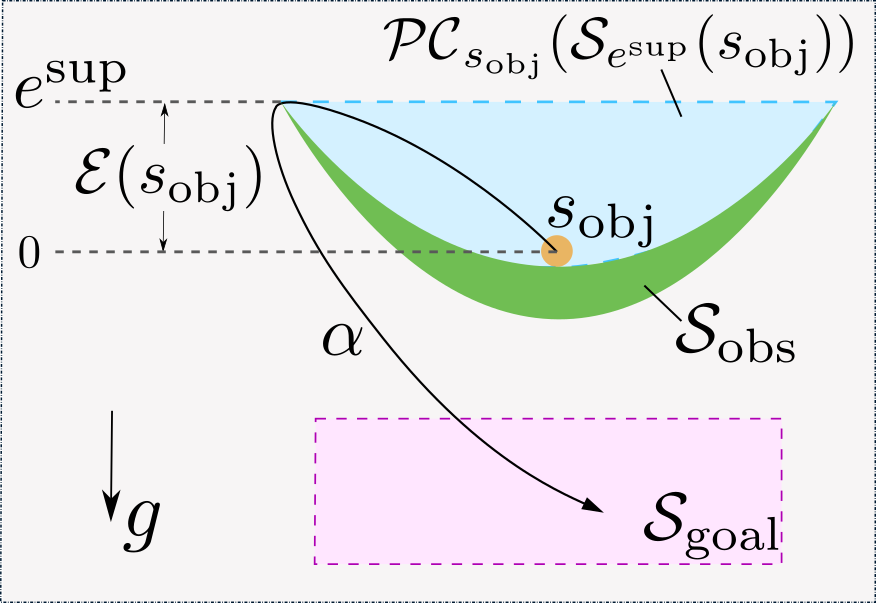}
    \caption{Illustration of energy-bounded caging under gravitational potential energy field $g$. The energy level set is distributed horizontally, such as $\mathcal{L}_{e^{\text{sup}}}(s_{\text{obj}})$. A point-mass object at configuration $s_{\text{obj}} \in \mathbb{R}^2$ lies inside a $\mathbb{R}^2$-bowl (green, $\mathcal{S}_{\text{obs}}$). The escape path $\alpha$ here has the minimum escape energy $\mathcal{E}(s_{\text{obj}})$.}
    \label{fig:ebc}
\end{figure}
}{}

\section{Caging-Guided Manipulator Co-Optimization}
\label{sec:4}

\subsection{Problem Formulation}
\label{sec:4-overview}
\begin{figure}[t]
    \centering
    \includegraphics[width=1\linewidth]{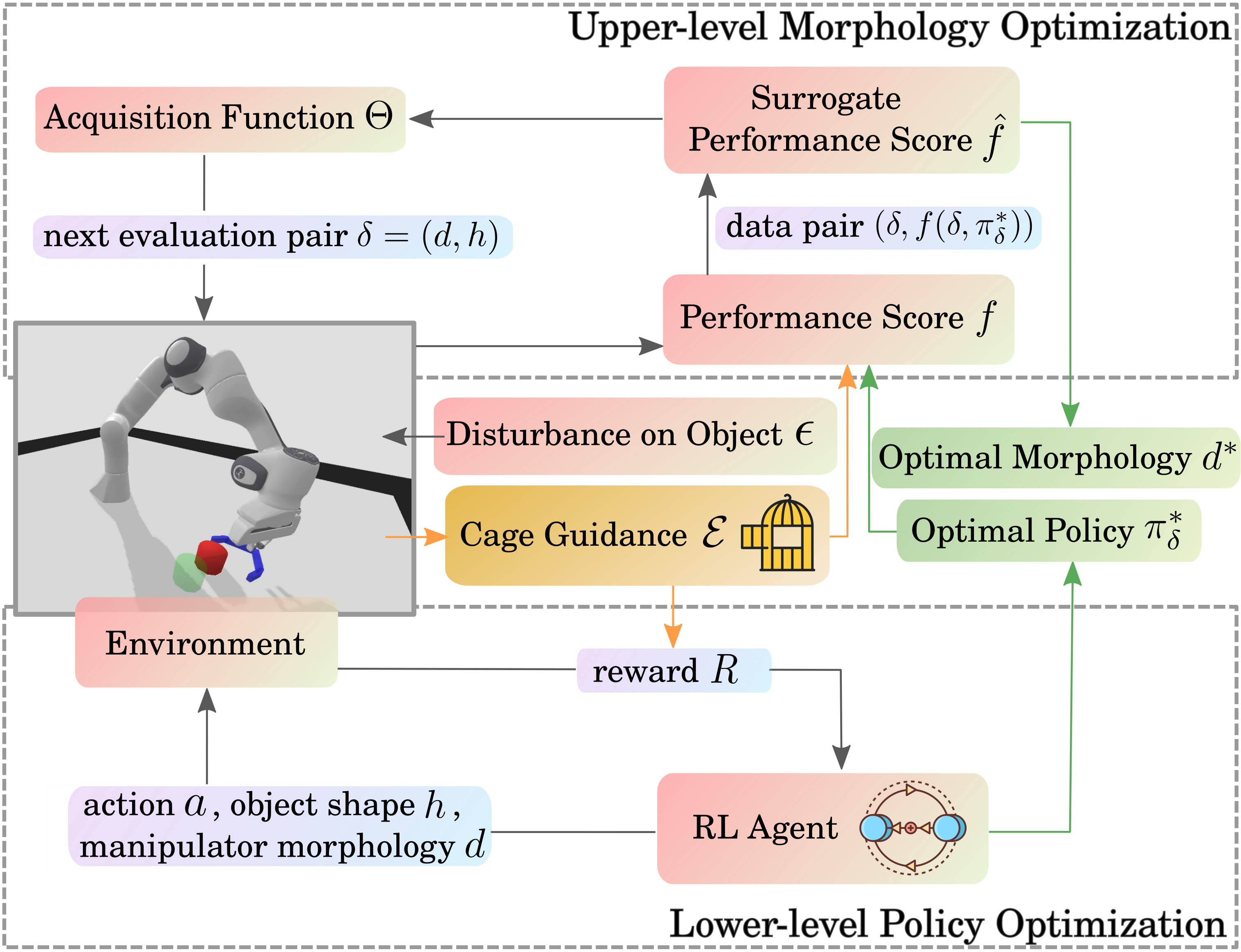}
    \caption{The proposed CageCoOpt framework.}
    \label{fig:pipeline}
\end{figure}

\yifei{Our framework co-optimizes the manipulator morphology $d$, and the control policy $\pi$, as illustrated in Fig. \ref{fig:pipeline}. $d$ in this paper refers to the morphology of robot end-effectors or rigid tools that robots use.  For a given $d \in \mathcal{D}$ in the morphology space $\mathcal{D}$, an optimal policy $\pi^*_d$ conditioned on $d$ is learned to maximize the expected cumulative reward $V(\pi | d)$. With the policy $\pi^*_d$, the morphology $d$ is evaluated by performing the manipulation task given $d$ and $\pi^*_d$ and obtaining a score $f(d, \pi^*_d)$. Iteratively, we select a different $d \in \mathcal{D}$ and repeat this process. Eventually, we obtain the optimal morphology $d^*$ with the highest performance score $f$. The co-optimization problem can be formulated as below,}
\begin{subequations}
\begin{align}
d^* &= \arg\max_d f(d, \pi^*_d) \label{eq:blo_1a} \\
\text{s.t.} \quad \pi^*_d &= \arg\max_{\pi} V(\pi | d) \label{eq:blo_1b},
\end{align}
\label{eq:blo1}
\end{subequations}
composed of the lower-level policy optimization (Eq.~\ref{eq:blo_1b}) and the upper-level morphology optimization (Eq.~\ref{eq:blo_1a}).
\yifei{Based on the formulation above, we consider a more general circumstance under uncertainties of object shape variations and unmodeled dynamics. Specifically, we consider the object shape space $\mathcal{H}$, and the manipulator-object pair $\delta = (d,h)$, $h \in \mathcal{H}$. The lower-level $V(\pi | \delta)$ is additionally conditioned on the object shape $h$ besides the morphology $d$. The upper-level score $f(\delta, \pi^*_\delta)$ is a function of the manipulator-object pair $\delta \in \mathcal{D} \times \mathcal{H}$ and the optimal policy $\pi^*_\delta$ conditioned on $\delta$. The optimization objective is the average score over all the possible object shapes $h$, i.e. $\mathbb{E}_{h \sim \mathcal{H}} f(\delta, \pi^*_\delta)$.
On the other hand, unmodeled dynamics exist in the manipulation scenario.
Following prior works~\cite{shirai2024robust, mahler2018dex, dong2024characterizing}, we treat it as random disturbance forces $\epsilon  \sim \mathcal{N}(0, \sigma_\epsilon^2)$ applied on the object's center of mass. The lower- and upper-level objectives are both conditioned on the randomized disturbance domain~\cite{andrychowicz2020learning}. Therefore, Eq.~\eqref{eq:blo1} can be reformulated as below,
}
\begin{subequations}
\begin{align}
d^* &= \arg\max_d \mathbb{E}_{h \sim \mathcal{H}} f(\delta, \pi^*_\delta | \epsilon) \label{eq:blo_2a} \\
\text{s.t.} \quad \pi^*_\delta &= \arg\max_{\pi} V(\pi | \delta, \epsilon) \label{eq:blo_2b}.
\end{align}
\label{eq:blo2}
\end{subequations}

This paper argues that integrating MEE into the co-optimization framework enhances manipulation robustness under the aforementioned uncertainties. We next discuss how we solve the bi-level optimization problem in Eq.~\eqref{eq:blo2}, and how we integrate MEE into the objectives. In the following, we use $d^*$, $\pi^*_\delta$ to denote the best-discovered solutions obtained from our stochastic optimization algorithms as an approximation of the global optimality.

\subsection{Lower-Level Policy Optimization}
\label{sec:4-rl}
A direct solution to the co-optimization in Eq.~\eqref{eq:blo2} often turns out inefficient, as training a new policy for each morphology $d\in\mathcal{D}$ requires extensive environment interaction. Moreover, training multiple specialists for similar tasks separately is redundant, as they can share knowledge. To address this, we decouple the bi-level problem by first learning a universal policy $\pi^*(s_t, \delta | \epsilon)$. This allows a single policy to generalize across different morphologies and object shapes, reducing training costs while maintaining adaptability. In other words, we solve the lower-level sub-problem Eq.~\eqref{eq:blo_2b} once to acquire a policy that adapts to any $(d,h)$. 

\yifei{Specifically, we formulate this policy optimization sub-problem as an infinite-horizon MDP $\big(\mathcal{S},\mathcal{A},\mathcal{P},\mathcal{R},\gamma\big)$. In each episode of training, a manipulator morphology $d \in \mathcal{D}$ and an object shape $h \in \mathcal{H}$ is selected at random (Line 4, Algo.~\ref{algo1}). At each timestep $t$, the agent observes a state $s_t\in\mathcal{S}$, and takes an action $a_t \in \mathcal{A}$. The environment transitions to $s_{t+1}$ through the transition dynamics $\mathcal{P}$ and provides a reward $r_t(s_t,\,a_t, \,s_{t+1} | \delta, \epsilon)$. The state space $\mathcal{S} = \mathcal{S}_{\text{obj}} \times \mathcal{S}_{\text{mnp}}$ is the composite configuration space of the object $\mathcal{S}_{\text{obj}}$ and the manipulator $\mathcal{S}_{\text{mnp}}$. The objective is to learn a policy $\pi^*_\delta$ that maximizes the expected cumulative reward $V$,
}
\begin{equation}
    V(\pi | \delta, \epsilon) = \mathbb{E}_{\pi} \left[\sum_{t=0}^{\infty} \gamma^t r_t(s_t,\,a_t,\,s_{t+1} | \delta, \epsilon)\right],
    \label{eq:r-R}
\end{equation}
\noindent where $\gamma$ denotes the discount factor. To enhance the manipulation robustness, we adopt a reward of the form:
\begin{equation}
    r_t = w_{\text{int}}r_{\text{int},t} + w_{\text{suc}}r_{\text{suc},t} + w_{\mathcal{E}}\mathcal{E}(s_{\text{obj}, t}),
    \label{eq:reward}
\end{equation}
\yifei{
where $r_{\text{int}}$ is a dense intermediate reward guiding progress toward the goal, $r_{\text{suc}}$ is a completion bonus, and $[w_{\text{int}}, w_{\text{suc}}, w_{\mathcal{E}}]$ is a set of weights. Crucially, we integrate the Minimum Escape Energy (MEE) $\mathcal{E}(s_{\text{obj},t})$, computed given the knowledge of the poses $s_t$ and the geometry $\delta$ of manipulator and object. It drives manipulator and object into \emph{energy-bounded caging} configurations. Such partial enclosure reduces the likelihood of object escape, even under random disturbances $\epsilon$. Furthermore, a caging configuration allows bounded mobility of the object inside it, thereby robust to the uncertainty of object shape variations $h \in \mathcal{H}$.
Additionally, we employ Proximal Policy Optimization (PPO)~\cite{schulman2017proximal} to train this universal policy $\pi^*_\delta$.
}

\subsection{Upper-Level Morphology Optimization}
\label{sec:4-bo}
\yifei{
The upper-level morphology optimization in Eq.~\eqref{eq:blo_2a} takes as input the universal policy $\pi^*_\delta$ from the lower level and aims to find the best manipulator morphology $d^* \in \mathcal{D}$. For each manipulator-object pair $\delta = (d,h)$, a policy rollout is performed to evaluate $\delta$ in the manipulation scenario. The computational burden increases as the dimensionality of $\mathcal{D}$ grows. Therefore, we aim at an algorithm with high sample efficiency rather than naive algorithms such as random search or grid search within $\mathcal{D}$. Inherently, the upper level is a derivative-free black-box optimization problem with costly evaluation function $f(\delta, \pi^*_\delta | \epsilon)$ and multi-dimensional input space $\mathcal{D} \times \mathcal{H}$. For this purpose, Bayesian optimization with a probabilistic surrogate model is a good fit. We employ Multi-Task Bayesian Optimization (MTBO)~\cite{swersky2013multi} to solve it, where the ``Multi-Task'' refers to multiple object shapes $h \in \mathcal{H}$. As a variant of BO, MTBO utilizes a surrogate function to approximate the true, computationally expensive one. Additionally, MTBO exploits correlations among the object shapes $h$, to further enhance data efficiency, allowing for more effective exploration of the input space $\mathcal{D} \times \mathcal{H}$.
}


Specifically, we define the \textit{performance score} $f$ as
\begin{align}
    f(\delta, \pi^*_\delta | \epsilon) = w\,\mathbb{E}_{s_0}[f_\text{suc}(\delta, \pi^*_\delta | \epsilon)] + (1-w)\,\mathbb{E}_{s_0}[f_\mathcal{E}(\delta, \pi^*_\delta | \epsilon)],
\label{eq:performance-score}
\end{align}
where $w \in [0,1]$ balances the success rate $f_\text{suc}$ and the robustness score $f_\mathcal{E}$. The robustness score $f_\mathcal{E}$ is a function of the average MEE~$\mathcal{E} (s_t, \delta)$ along a rollout trajectory $\theta: [0,T] \rightarrow \mathcal{S}$,
\begin{align}
    f_\mathcal{E}(\delta, \pi^*_\delta | \epsilon) = \frac{1}{T}\sum_{t=0}^T \mathcal{E} (s_{\text{obj},t}).
\label{eq:upper-cage}
\end{align}

\begin{algorithm}[tp]
\SetAlgoLined
\caption{Caging-Guided Manipulator Morphology and Policy Co-Optimization}
\label{algo1}
\small

\textbf{Lower-Level Policy Optimization}\\
\KwIn{Morphology space $\mathcal{D}$, object shape space $\mathcal{H}$, initial policy $\pi_0$, maximum iterations $N_{\text{rl}}$.}
\For{$n = 0$ to $N_{\text{rl}}$}
{
    Randomize initial object and robot state $s_0$ \\
    Randomize morphology $d_n \in \mathcal{D}$ and object shape $h_n \in \mathcal{H}$\\
    Run policy $\pi_n(s_t, \delta_n)$ in environment for $N_{\text{step}}$ time steps, where $\delta_n = (d_n, h_n)$\\
    Update policy $\pi_{n+1}$ via PPO \\
}
\BlankLine
\textbf{Upper-Level Morphology Optimization}\\
\KwIn{Best universal policy $\pi^*_{\delta}$, initial morphology $d_0 \in \mathcal{D}$, object shape $h_0 \in \mathcal{H}$, $\delta_0 = (d_0, h_0)$, maximum iterations $N_{\text{bo}}$.}

\For{$i = 0$ to $N_{\text{bo}}$}
{
    Compute next evaluation pair $\delta_{i+1} \gets \arg\max_\delta \Theta_{i}(\delta, \hat{f}_i)$ \hfill {\footnotesize $\triangleright$ \textcolor{blue}{\textsc{eq. \eqref{eq:acq-func}, \eqref{eq:next-query}}}}  \\
    Evaluate performance score $f(\delta_{i+1}, \pi^*_{\delta_{i+1}} | \epsilon)$ \hfill {\footnotesize $\triangleright$ \textcolor{blue}{\textsc{eq. \eqref{eq:performance-score}}}}\\
    Update GP mean and kernel functions $\hat{f}_{\mu,i+1}$, $\hat{f}_{k,i+1}$ with $(\delta_{i+1}, f(\delta_{i+1}, \pi^*_{\delta_{i+1}} | \epsilon))$ \\
}
Compute best-found morphology $d^*$ \hfill {\footnotesize $\triangleright$ \textcolor{blue}{\textsc{eq. \eqref{eq:optimal-morph}}}}  \\

\Return Best policy $\pi^*_{\delta}$, best morphology $d^*$
\end{algorithm}

By integrating MEE into the performance score $f$, the morphology $d$ that favors energy-bounded caging configurations is selected. Additionally, random perturbation forces $\epsilon \sim \mathcal{N}(0, \sigma_\epsilon^2)$ applied on the object persist throughout the evaluation rollouts, similar to the lower level. The expectations in Eq.~\eqref{eq:performance-score} are approximately evaluated as the mean cumulative scores for a fixed number of policy rollouts given random initial states $s_0 \in \mathcal{S}$. 

We employ the Gaussian Process (GP) as the \textit{surrogate performance score} $\hat{f}(\delta, \pi^*_\delta | \epsilon)$. It maps $\delta$ to its predicted mean performance score $f_{\mu}$ and the corresponding kernel function $f_{k}$, i.e. ${\hat{f}: \delta \mapsto (\hat{f}_{\mu}(\delta, \pi^*_\delta | \epsilon), \hat{f}_{k}(\delta, \pi^*_\delta | \epsilon))}$. Specifically, we use the intrinsic co-regionalization model~\cite{bonilla2007multi} as the kernel. It improves sample efficiency by modeling correlations between morphology $d \in \mathcal{D}$ and object shapes $h \in \mathcal{H}$. 

At iteration $i$, an Upper Confidence Bound (UCB)-based~\cite{garivier2011upper}  \textit{acquisition score function} is employed to predict the utility of sampling a particular $\delta$ based on the current surrogate model $\hat{f}_{i}$:
\begin{align}
\Theta_{i}(\delta, \hat{f}_{i}) = \hat{f}_{\mu, {i}}(\delta, \pi^*_\delta | \epsilon) + \lambda \cdot \hat{f}_{k, {i}}(\delta, \pi^*_\delta | \epsilon).
\label{eq:acq-func}
\end{align}
The weight $\lambda$ adjusts the exploration-exploitation trade-off. The acquisition score function is then used as 
\begin{align}
    \delta_{i+1} = \argmax_\delta \Theta_{i}(\delta, \hat{f}_{i})
\label{eq:next-query}
\end{align}
to determine the next best query input $\delta_{i+1}$ (Line 10, Algo.~\ref{algo1}). We then evaluate the performance score~$f(\delta_{i+1}, \pi^*_{\delta_{i+1}} | \epsilon)$ of this query by rolling out with policy $\pi^*_{\delta_{i+1}}$ (Line 11) and thereby obtain a new data pair ${(\delta_{i+1}, f(\delta_{i+1}, \pi^*_{\delta_{i+1}} | \epsilon))}$. The data pair is used for updating the surrogate mean and kernel functions ${(f_{\mu, {i+1}}, f_{k, {i+1}})}$ (Line 12).
Given the last fit of the surrogate $\hat{f}_{N_{\text{bo}}}$ at iteration $N_{\text{bo}}$, we determine the best-found morphology $d^*$ (Line 14) by
\begin{equation}
    d^* = \argmax_d \mathbb{E}_{h \sim \mathcal{H}} \left[ \hat{f}_{\mu, N_{\text{bo}}}(\delta, \pi^*_\delta | \epsilon) \right].
\label{eq:optimal-morph}
\end{equation}

Together with the lower level, the CageCoOpt framework outputs the best-discovered policy $\pi^*_{\delta}$ and morphology $d^*$ for caging-based manipulation. 

\ifthenelse{\boolean{includeFigures}}{
\begin{figure}[t]
    \centering
    \includegraphics[width=0.95\linewidth]{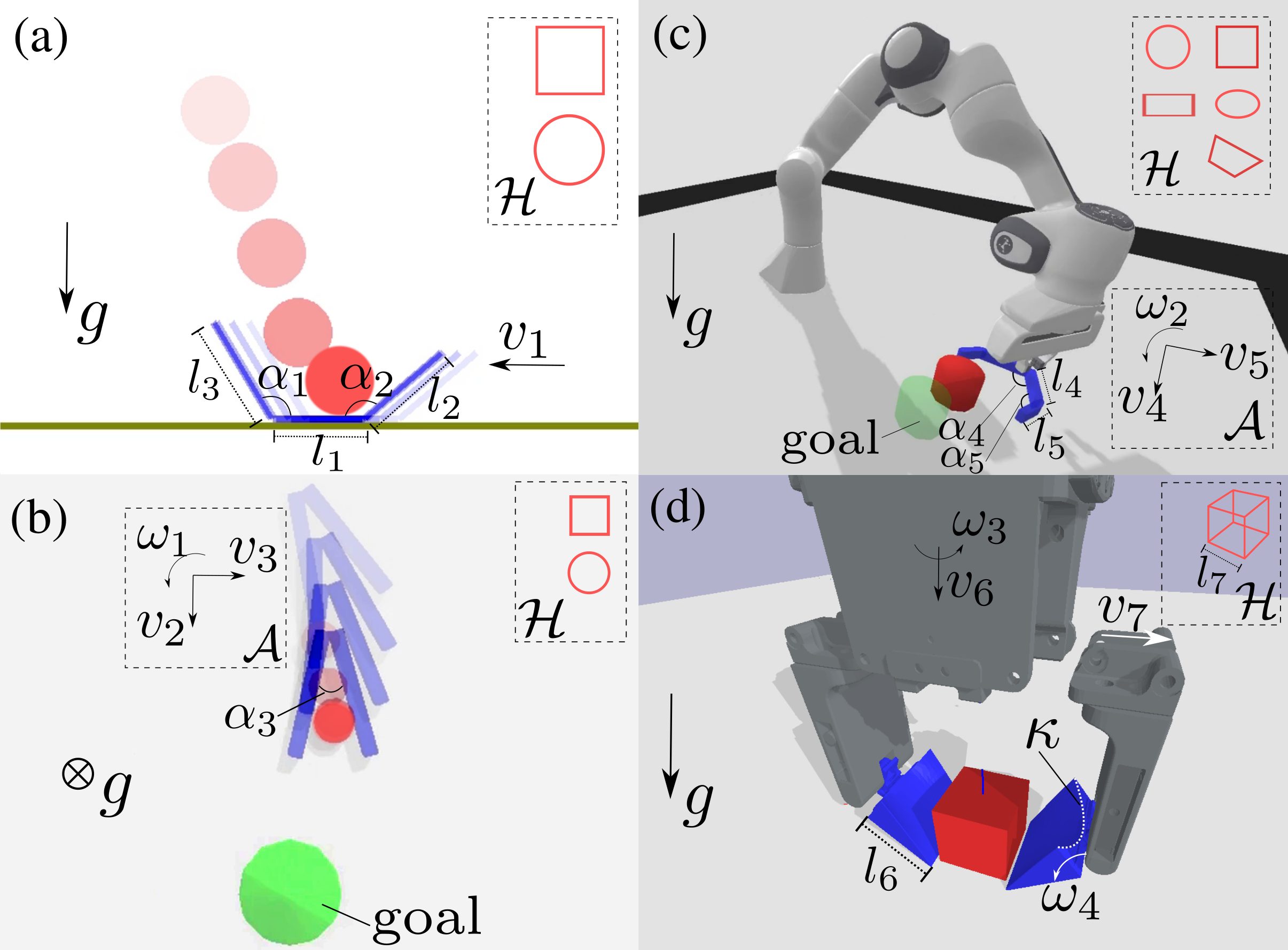}
    \caption{Four manipulation tasks detailed in Section \ref{sec5a}. Gravity is denoted by $g$.}
    \label{fig:envs}
\end{figure}
}{}

\ifthenelse{\boolean{includeFigures}}{
\begin{figure}[t]
    \centering
    \includegraphics[width=\linewidth]{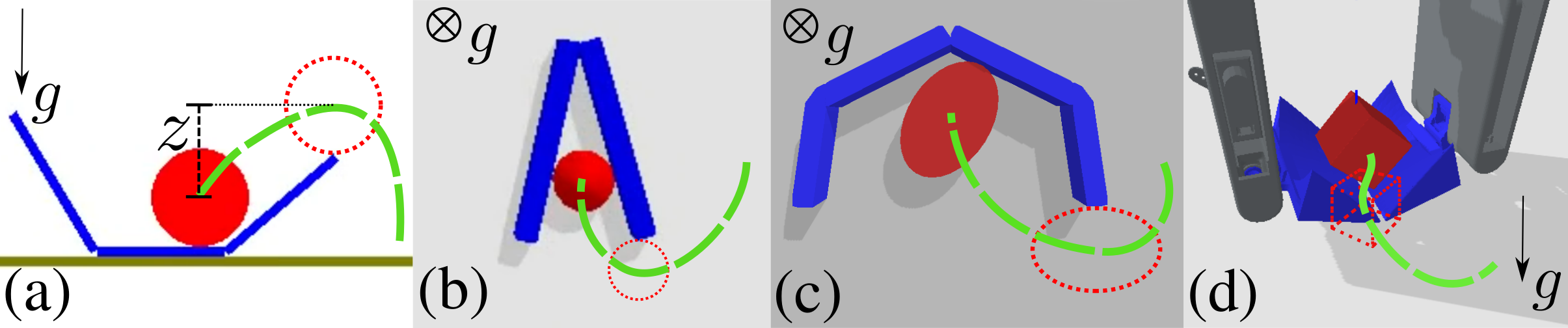}
    \caption{Escape paths (green) with minimum escape energy $\mathcal{E}(s_{\text{obj}})$.}
    \label{fig:cage}
\end{figure}
}{}


\section{Evaluation}
\yifei{We design four manipulation tasks (Fig.~\ref{fig:envs}) to answer several research questions and demonstrate the efficacy of our proposed approach\footnote{For more details, visit \url{https://sites.google.com/view/robust-codesign/}.}.}

\subsection{Manipulation Task}
\label{sec5a}
The morphology spaces $\mathcal{D}$ of the manipulation tasks range from 1 to 5 dimensions. The action spaces $\mathcal{A}$ range from 1 to 6 dimensions.
\yifei{We consider relatively low-dimensional morphology space $\mathcal{D}$, such as linkage representations of manipulators, following prior co-design works~\cite{liu2023learning, liu2024paperbot}. The simple and low-dimensional linkage morphology space captures the core functionality of manipulators, and also alleviates the computational challenge for bi-level optimization problems with high-dimensional morphology space.}
Pybullet~\cite{coumans2016pybullet} and Box2D \cite{catto_box2d} are used for physics simulation. 

\texttt{Catch}: The goal is to design a basket to catch a falling object under gravity (Fig.~\ref{fig:envs}-a). The manipulator morphology space is ${\mathcal{D}=\{l_1,l_2,l_3,\alpha_1,\alpha_2\}}$, specifying the linkage lengths and angles. The action space includes the basket’s horizontal velocity $v_1$. The object shape space is $\mathcal{H} = \{\texttt{circle}, \texttt{square}\}$.

\texttt{VPush}: The goal is to design a symmetric V-shaped tool to push an object into a circular goal region (Fig.~\ref{fig:envs}-b). The morphology space includes the opening angle $\alpha_3$. The action space $\mathcal{A}$ is $\{v_2,v_3,\omega_1\}$, and ${\mathcal{H}=\{\texttt{circle},\texttt{square}\}}$.

\texttt{Panda-UPush}: The goal is to design a U-shaped manipulator mounted on a Franka arm to push an object into a circular goal region (Fig.~\ref{fig:envs}-c). The morphology space is $\mathcal{D}=\{\alpha_4,\alpha_5,l_4,l_5\}$. The action space contains $\{v_4,v_5,\omega_2\}$. The object shape space is $\mathcal{H} = \{\texttt{circle}, \texttt{rectangle}, \texttt{square}, \texttt{oval}, \texttt{irregular}\}$.

\texttt{Scoop}: In Fig.~\ref{fig:envs}-d, we have a linear scoop gripper with variable stiffness for adaptive gripping. It employs a double parallelogram mechanism (hidden in Fig.~\ref{fig:envs}-d) on each side to ensure parallel finger closing. 
The goal is to design the scoop gripper tips for robustly lifting a cube by at least 0.06 m within 1.0 s. The morphology space $\mathcal{D}$ includes the fingertip length $l_6$ and the curvature~$\kappa$. The action space is $\{v_6,\omega_3,v_7,\omega_4\}$. The continuous shape space contains $l_7\in[0.01,0.02]$.

Fig.~\ref{fig:cage} illustrates robust object poses in energy-bounded caging configurations, exhibiting positive Minimum Escape Energy (MEE) values. The most energy-efficient escape paths are highlighted, corresponding to the MEE, where objects navigate closely around the manipulator edges to minimize the work required against gravitational forces~(a,~d) or frictional resistance (b, c). In practice, MEE in some tasks is computed analytically. For instance, in \texttt{Catch}, the minimum escape energy against the gravity is $\mathcal{E}(s_{\text{obj}})=m_{\text{obj}}gz$. $m_{\text{obj}}$ is the mass of the object and $z$ denotes the height difference in Fig.~\ref{fig:envs}-a.

\ifthenelse{\boolean{includeFigures}}{
\begin{figure*}[!t]
    \centering
    \input{assets/tex/rob_experiment/envs_success_rate}
    \caption{Training success rate w.r.t. training steps in \texttt{Catch}, \texttt{VPush}, \texttt{Panda-UPush}, and \texttt{Scoop}, respectively. We include (w/) or exclude (w/o) the MEE metric in the RL reward and the disturbances $\epsilon$ on the object. Each curve represents the mean over $N_{\text{seed}}=5$ random seeds. The standard deviation shades for curves without training disturbances (w/o dist.) are omitted for visual clarity. 
    }
    \label{fig:rl-train}
\end{figure*}
}{}

\begin{table*}[t]
\centering
\caption{Comparison of MTBO with baselines in morphology optimization.}
\footnotesize
\setlength{\tabcolsep}{3pt}
\begin{tabular}{cc cccc c}
\toprule
Method & MEE & \texttt{Catch} & \texttt{VPush} & \texttt{Panda-UPush} & \texttt{Scoop} & $\Bar{N}_{\text{roll}}$ \\
\midrule
\multirow{2}{*}{MTBO (ours)} 
 & w/  & $0.91 \pm \mathbf{0.06}$ & $\mathbf{0.85} \pm \mathbf{0.03}$ & $\mathbf{0.53} \pm 0.20$ & $\mathbf{0.93} \pm 0.11$ & \multirow{2}{*}{$\mathbf{650}$} \\
 & w/o & $0.67 \pm 0.17$ & $0.80 \pm 0.07$ & $0.23 \pm 0.17$ & $0.71 \pm 0.36$ &  \\
\midrule
\multirow{2}{*}{BO} 
 & w/  & $\mathbf{0.92} \pm 0.08$ & $0.83 \pm 0.09$ & $0.44 \pm \mathbf{0.11}$ & $0.90 \pm 0.07$ & \multirow{2}{*}{$690$} \\
 & w/o & $0.75 \pm 0.13$ & $0.83 \pm 0.14$ & $0.22 \pm 0.20$ & $0.77 \pm 0.39$ &  \\
\midrule
\multirow{2}{*}{GA} 
 & w/  & $0.91 \pm 0.07$ & $0.79 \pm 0.12$ & $0.53 \pm 0.15$ & $0.91 \pm \mathbf{0.10}$ & \multirow{2}{*}{$1140$} \\
 & w/o & $0.69 \pm 0.11$ & $0.82 \pm 0.10$ & $0.26 \pm 0.21$ & $0.73 \pm 0.38$ & \\
\bottomrule
\end{tabular}
\label{tab:baselines}
\end{table*} 

\subsection{Experiments}
\label{sec5b}

\textit{1. Does the incorporation of MEE improve performance under uncertainty?}

We demonstrate that integrating the caging-based robustness metric, MEE $\mathcal{E}(s_\text{obj})$, in the objective functions does improve the performance both at the lower-level policy learning and upper-level morphology optimization, especially in the presence of unmodeled dynamics. We add random disturbance forces $\epsilon  \sim \mathcal{N}(0, \sigma_\epsilon^2)$ on the object to simulate unmodeled dynamics. The success rate serves as the primary performance metric, where success is defined as follows: in \texttt{catch}, the object falls into the basket; in \texttt{VPush} and \texttt{Panda-UPush}, the object reaches the goal region; and in \texttt{Scoop}, the object is lifted.

\yifei{\textit{Policy Optimization}: The results of RL policy learning are shown in Fig.~\ref{fig:rl-train}. Incorporating MEE into the RL reward (Eq.~\ref{eq:reward}) consistently improves policy performance, particularly under unmodeled dynamics (w/ dist.). The inclusion of MEE encourages caging configurations, where the manipulator partially cages the object, forming energy barriers that prevent escapes. As a result, disturbances with energy levels below the minimum escape energy cannot devastate the energy-bounded cage. Moreover, MEE-based policies exhibit robustness to object geometric uncertainties, as caging relies on relaxed geometric constraints rather than precise contact conditions. This leads to higher manipulation success rates. The performance in \texttt{Panda-UPush} is lower than in \texttt{VPush} due to the increased complexity of the morphology space $\mathcal{D}$ and the stricter goal completion criteria.}

\textit{Morphology Optimization}: The first two rows of Table~\ref{tab:baselines} demonstrate the impact of integrating MEE into the performance score $f$ (Eq.~\ref{eq:performance-score}) at the upper-level morphology optimization.
The data in Table~\ref{tab:baselines} represent the confidence threshold of the test success rate $Q$. The mean $Q_{\mu}$ is computed using the best-found morphologies $d^*_i$ from the last five iterations:
\begin{equation}
    Q_{\mu} = \sum^{N_{\text{seed}}}_{j=1} \sum^{N_{\text{bo}}}_{i=N_\text{bo}-4}{\mathbb{E}_{h \sim \mathcal{H}}[f_\text{suc}((d^*_i, h), \pi^*_{\delta, j} | \epsilon)]}.
\end{equation}
The evaluation policies $\pi^*_{\delta, j}$ are from the five randomly seeded models trained at the lower-level RL. MEE is either incorporated at both levels or excluded entirely. Rollouts are conducted with the object subjected to unmodeled dynamics (w/ dist.). The metric $\Bar{N}_{\text{roll}}$ denotes the minimum average number of control policy rollouts required for convergence in morphology optimization.

\yifei{Results indicate that integrating MEE into the MTBO performance score $f$ consistently improves success rates, highlighting the advantages of caging-based manipulator morphology for robust manipulation. Fig.~\ref{fig:qualitative} illustrates the identified morphologies throughout the optimization process, The results reveal a clear preference for morphologies that partially cage the object and prevent its effortless escape. This caging-based strategy significantly enhances manipulation success across diverse object shapes and in the presence of unmodeled dynamics.}

\ifthenelse{\boolean{includeFigures}}{
\begin{figure}[!t]
    \centering
    \input{assets/tex/perturb_experiment/perturb_plot}  
    \caption{Success rate across varying disturbance intensity $\sigma_\epsilon$ in testing environments (From left to right: \texttt{Catch}, \texttt{VPush}, \texttt{Scoop}). w/ or w/o dist. means the inclusion or exclusion of random disturbance forces on the objects during training, respectively. Each data point represents the average of 250 rollouts (50 rollouts per seed, across 5 random seeds corresponding to the 5 RL models from Fig. \ref{fig:rl-train}). }
    \label{fig:disturbance}
\end{figure}
}{}

\ifthenelse{\boolean{includeFigures}}{
\begin{figure}[t]
    \centering
    \includegraphics[width=\linewidth]{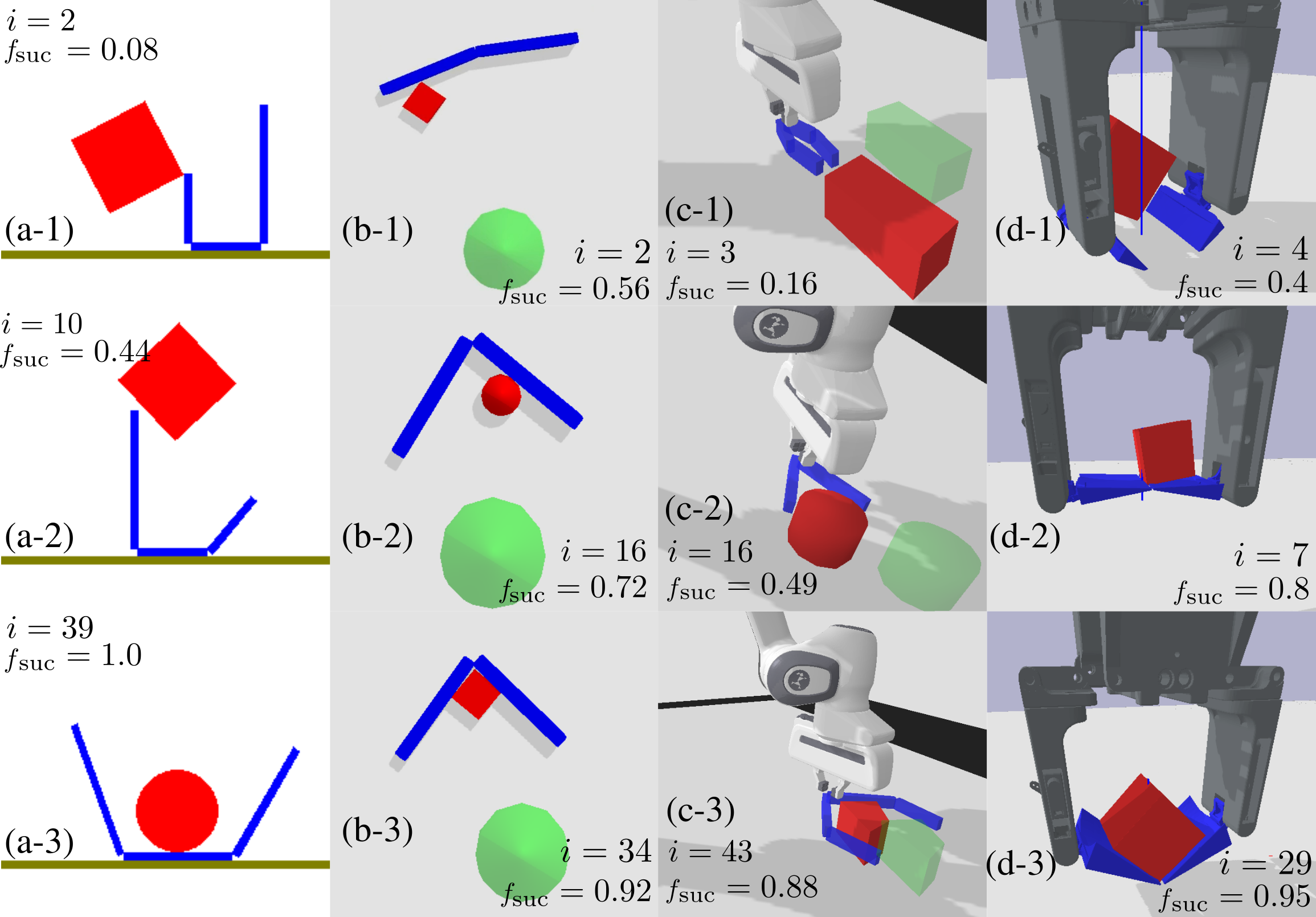}
    \caption{Morphology optimization process. $i$ denotes the iteration number, and $f_{\text{suc}}$ the test success rate given the manipulator morphology $d$. Here, the optimizer is MTBO with MEE. }
    \label{fig:qualitative}
\end{figure}
}{}

\textit{2. How does unmodeled dynamics affect performance?}

\yifei{
We conduct an ablation study to evaluate how unmodeled dynamics, simulated as random disturbance forces $\epsilon \sim \mathcal{N}(0, \sigma_\epsilon^2)$, influence RL model performance (Fig.~\ref{fig:disturbance}). The results indicate a general decline in success rate as disturbance intensity $\sigma_\epsilon$ increases. Models trained with MEE consistently outperform those without it, regardless of whether disturbances are present during training. Moreover, when disturbances are introduced during training (w/ dist.), MEE provides a greater performance boost compared to training in a disturbance-free environment. This suggests that the caging-based manipulation strategy, reinforced by the MEE robustness metric, enhances the manipulator’s ability to cope with uncertainty.
}

\textit{3. Is MTBO more sample-efficient than the baselines?}

\yifei{
We compare MTBO with two baselines: standard BO and 
Genetic Algorithm (GA) \cite{kramer2017genetic}. For BO, we use a Lower Confidence Bound acquisition function~\cite{srinivas2009gaussian}. For GA, we set a population size of 4 and a mutation rate of 0.1. In both methods, morphology $d$ is evaluated across all object shapes $h \in \mathcal{H}$, with success rates averaged over multiple rollouts.  Unlike BO and GA, MTBO selects a specific $\delta = (d, h)$ in each iteration and optimizes by leveraging correlations between object shapes $h$. This approach reduces the number of required rollouts while maintaining high performance. MTBO achieves better results ($Q_\mu \pm Q_\sigma$) with fewer rollouts $\Bar{N}_{\text{roll}}$ in most environments (Table \ref{tab:baselines}), demonstrating better sample efficiency. We attribute this improvement to MTBO’s use of Gaussian Processes (GP) to model correlations between object shapes $h$, allowing for more effective exploration. This strategy enhances manipulation robustness under the uncertainty of object shapes without significantly increasing computational cost.
}

\textit{4. Do caging-guided co-optimized manipulators improve real-world robustness?}

We conduct a proof-of-concept physical evaluation with a Franka Emika Panda arm, attaching three manipulators from our upper-level morphology optimization (Fig.\ref{fig:qualitative}-c, \texttt{Panda-UPush}). We chose two representative objects, \texttt{circle} and \texttt{square} from $\mathcal{H}$, and tracked their poses via a motion-capture system. Uncertainties primarily arise from variations in object shape, inaccuracies in contact modeling, and positioning errors in the motion capture system.

The results, shown in Fig.~\ref{fig:real-world}, indicate qualitatively that the learned policies can transfer from simulation to the physical robot. Specifically, we observe that the initially narrow, ``chopstick-like'' manipulator (Fig.\ref{fig:real-world}, a-1) frequently fails to poke the circular object to the goal due to slippage at contact, consistent with our failure cases in simulation (Fig.\ref{fig:teaser}-a). In contrast, the co-optimized manipulators (Fig.~\ref{fig:real-world}, b-1,c-1) more reliably drive both objects to the goal. The interaction strategy shifts from poking the object with the manipulator's distal ends to using the inner surface for continuous nudging, thereby maintaining a partial cage.
The object slides deep within the partial cage, allowing the manipulator to push it ``blindly'' while the object naturally follows without requiring precise contacts. With finite RL training steps in practice, the learned universal policy $\pi^*_\delta$ is often suboptimal. Under this condition, the caging-based strategy (Fig.\ref{fig:teaser}-b) outperforms the contact-dependent one (Fig.\ref{fig:teaser}-a), as it is less reliant on control policies with high contact accuracy.  
These results support our simulation findings: co-optimized morphology and policy enhance robustness against real-world uncertainties through caging strategies.

\ifthenelse{\boolean{includeFigures}}{
\begin{figure}[t]
    \centering
    \includegraphics[width=0.75\linewidth]{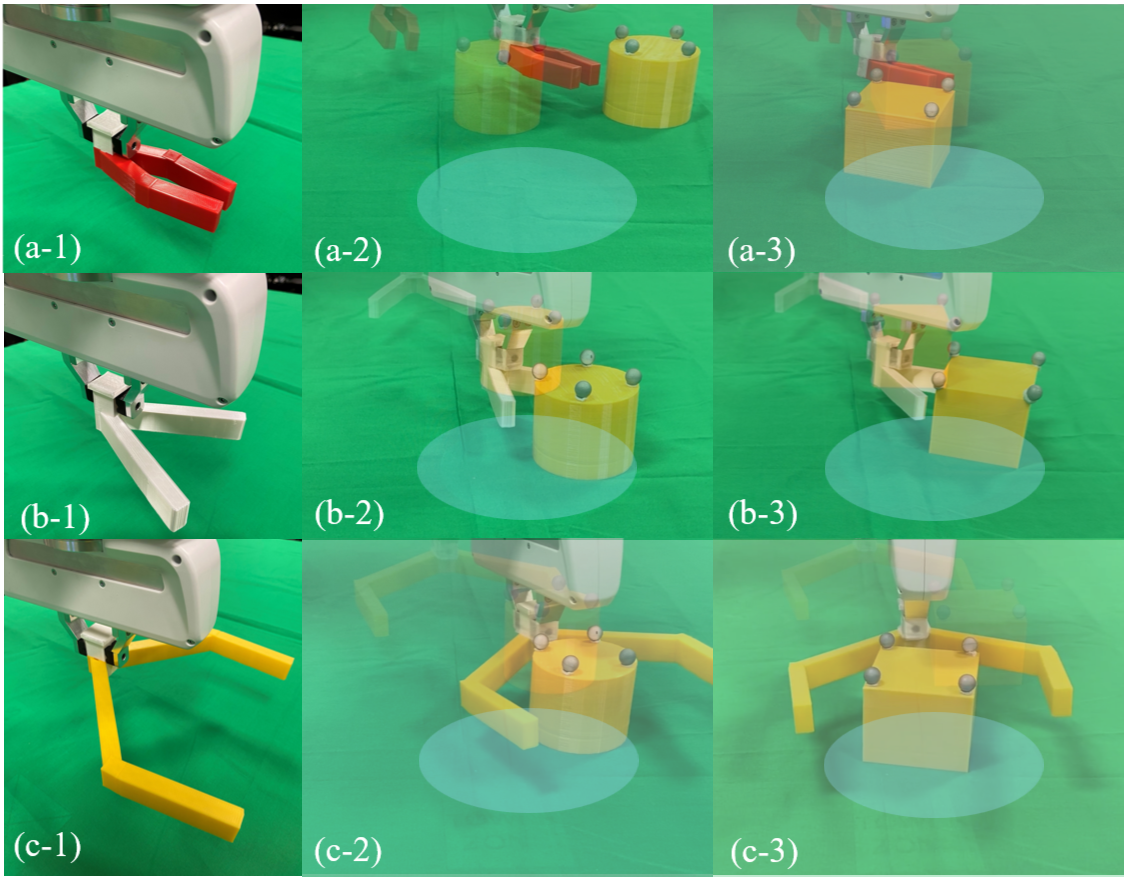}
    \caption{Real-world experiments. The goal regions are shown in blue.}
    \label{fig:real-world}
\end{figure}
}{}

\section{Conclusion}
\yifei{In this work, we explored the application of caging, a concept from classic grasp theory, to data-driven manipulator morphology and policy co-optimization for robust manipulation. We demonstrated the effectiveness of the caging-based robustness metric, Minimum Escape Energy, in designing and controlling manipulators under object geometric uncertainty and unmodeled dynamics. A limitation is that the current evaluation still focuses on simple morphology spaces, leading to morphologies aligning with human instinct. For future work, we believe extending this approach to complex morphology spaces will potentially allow the discovery of novel and even counter-intuitive robot morphology for robust manipulation. Future work on real-world experiments can incorporate real-time impedance adaptation with smoother executions and provide quantitative evaluations. Additionally, we aim to explore the capabilities of this framework further in co-optimizing manipulators for robust manipulation of deformable objects.}

\section*{Acknowledgment}
Thanks to Mengyuan Zhao, Kei Ikemura and Alon Shirizly for proofreading and Zahid Muhammad and Haofei Lu for assistance with the experiments.


\balance
\bibliographystyle{bibliography/IEEEtran}
\bibliography{bibliography/main}

\ifthenelse{\boolean{includeAppendix}}{
\clearpage
\section{Appendix}
}{}

\end{document}

%% file: preamble.tex
\usepackage{pifont}
\usepackage{todonotes}

\def\M{\mathcal{M}}

\usepackage[nolist]{acronym} 

\newacro{dof}[DoF]{Degree of Freedom}
\acrodefplural{dof}{Degrees of Freedom}
\newacro{mujoco}[MuJoCo]{MuJoCo}
\newacro{bullet}[Bullet]{Bullet}
\newacro{sofa}[SOFA]{SOFA}
\newacro{flex}[Flex]{Flex}
\newacro{isaacsim}[IsaacSim]{IsaacSim}
\newacro{physx}[PhysX]{PhysX}
\newacro{dcm}[DCM]{Dynamic Cloth Manipulation}
\newacro{fem}[FEM]{Finite Element Method}
\newacro{pbd}[PBD]{Position Based Dynamics}
\newacro{cd}[CD]{Chamfer Distance}
\newacro{hd}[HD]{Hausdorff Distance}
\newacro{point_cloud}[point cloud]{point cloud}
\newacro{timestep}[time step]{time step}
\newacro{timepoint}[time point]{time point}
\newacro{data_set}[dataset]{dataset}
\newacro{rigid_object}[rigid-object]{rigid-object}
\newacro{set_up}[set-up]{set-up}
\newacro{cmaes}[CMA-ES]{Covariance Matrix Adaptation Evolution Strategy}
\newacro{bo}[BO]{Bayesian Optimisation}

\usepackage{xcolor,colortbl}

\definecolor{lavender}{rgb}{0.9, 0.9, 0.98}
\definecolor{red}{RGB}{255, 0, 0}
\definecolor{orange}{RGB}{252, 130, 62}
\definecolor{blue}{RGB}{0, 0,255}
\definecolor{darkgreen}{RGB}{0, 150,0}
\newcommand{\note}[1]{\color{orange}}



%% file: assets/tex/perturb_experiment/perturb_plot.tex
\begin{tikzpicture}

\definecolor{darkgray176}{RGB}{176,176,176}
\definecolor{darkorange25512714}{RGB}{255,127,14}
\definecolor{steelblue31119180}{RGB}{31,119,180}

\begin{groupplot}[group style={group size=3 by 1, horizontal sep=0.1cm}]
\nextgroupplot[
height=5cm,
tick align=inside,
tick pos=left,
width=4cm,
x grid style={darkgray176},
xlabel={},
xmin=0, xmax=0.9,
xtick style={color=black},
y grid style={darkgray176},
ylabel={\footnotesize Testing Success Rate},
ymin=0, ymax=1,
ytick style={color=black},
]
\path [draw=steelblue31119180, fill=steelblue31119180, opacity=0.2]
(axis cs:0,1.02330302779823)
--(axis cs:0,0.656696972201766)
--(axis cs:0.1,0.668489554621465)
--(axis cs:0.2,0.662572328461174)
--(axis cs:0.3,0.633589916233409)
--(axis cs:0.4,0.57287684822792)
--(axis cs:0.5,0.622257903387003)
--(axis cs:0.6,0.546458434959374)
--(axis cs:0.7,0.470871215252208)
--(axis cs:0.8,0.490531598666243)
--(axis cs:0.9,0.418479474571144)
--(axis cs:0.9,0.893520525428856)
--(axis cs:0.9,0.893520525428856)
--(axis cs:0.8,0.941468401333757)
--(axis cs:0.7,0.929128784747792)
--(axis cs:0.6,0.973541565040626)
--(axis cs:0.5,1.009742096613)
--(axis cs:0.4,0.98712315177208)
--(axis cs:0.3,1.01441008376659)
--(axis cs:0.2,1.02542767153883)
--(axis cs:0.1,1.02751044537854)
--(axis cs:0,1.02330302779823)
--cycle;

\path [draw=darkorange25512714, fill=darkorange25512714, opacity=0.2]
(axis cs:0,0.86619000805153)
--(axis cs:0,0.38180999194847)
--(axis cs:0.1,0.324605578074201)
--(axis cs:0.2,0.316056458039335)
--(axis cs:0.3,0.320322790713396)
--(axis cs:0.4,0.195572948332916)
--(axis cs:0.5,0.262072010370987)
--(axis cs:0.6,0.173220746414939)
--(axis cs:0.7,0.218512525364498)
--(axis cs:0.8,0.151467793532222)
--(axis cs:0.9,0.106479474571144)
--(axis cs:0.9,0.581520525428856)
--(axis cs:0.9,0.581520525428856)
--(axis cs:0.8,0.640532206467778)
--(axis cs:0.7,0.717487474635502)
--(axis cs:0.6,0.666779253585061)
--(axis cs:0.5,0.761927989629013)
--(axis cs:0.4,0.692427051667084)
--(axis cs:0.3,0.815677209286603)
--(axis cs:0.2,0.811943541960665)
--(axis cs:0.1,0.819394421925798)
--(axis cs:0,0.86619000805153)
--cycle;

\addplot [semithick, steelblue31119180]
table {%
0 0.84
0.1 0.848
0.2 0.844
0.3 0.824
0.4 0.78
0.5 0.816
0.6 0.76
0.7 0.7
0.8 0.716
0.9 0.656
};

\addplot [semithick, darkorange25512714]
table {%
0 0.624
0.1 0.572
0.2 0.564
0.3 0.568
0.4 0.444
0.5 0.512
0.6 0.42
0.7 0.468
0.8 0.396
0.9 0.344
};

\addplot [semithick, steelblue31119180, dashed]
table {%
0 0.9
0.1 0.892
0.2 0.888
0.3 0.896
0.4 0.824
0.5 0.788
0.6 0.784
0.7 0.72
0.8 0.7
0.9 0.652
};
\addplot [semithick, darkorange25512714, dashed]
table {%
0 0.832
0.1 0.796
0.2 0.804
0.3 0.76
0.4 0.708
0.5 0.704
0.6 0.656
0.7 0.564
0.8 0.564
0.9 0.496
};

\nextgroupplot[
height=5cm,
tick align=inside,
tick pos=left,
width=4cm,
x grid style={darkgray176},
xlabel={\footnotesize Disturbance Intensity $\sigma_\epsilon$},
xmin=0, xmax=0.9,
xtick style={color=black},
y grid style={darkgray176},
ymin=0, ymax=1,
ytick style={color=black},
ytick=\empty,
legend style={font=\tiny, at={(0.14,0.01)}, anchor=south west}, 
legend cell align={left},
legend image post style={xscale=0.4},
]
\path [draw=steelblue31119180, fill=steelblue31119180, opacity=0.2]
(axis cs:0,0.97631109250343)
--(axis cs:0,0.55168890749657)
--(axis cs:0.1,0.485584452830641)
--(axis cs:0.2,0.515600362976705)
--(axis cs:0.3,0.505505056237226)
--(axis cs:0.4,0.470871215252208)
--(axis cs:0.5,0.495500556793564)
--(axis cs:0.6,0.475754518117006)
--(axis cs:0.7,0.432534503589167)
--(axis cs:0.8,0.359467793532222)
--(axis cs:0.9,0.38180999194847)
--(axis cs:0.9,0.86619000805153)
--(axis cs:0.9,0.86619000805153)
--(axis cs:0.8,0.848532206467778)
--(axis cs:0.7,0.903465496410833)
--(axis cs:0.6,0.932245481882994)
--(axis cs:0.5,0.944499443206436)
--(axis cs:0.4,0.929128784747792)
--(axis cs:0.3,0.950494943762774)
--(axis cs:0.2,0.956399637023295)
--(axis cs:0.1,0.938415547169358)
--(axis cs:0,0.97631109250343)
--cycle;

\path [draw=darkorange25512714, fill=darkorange25512714, opacity=0.2]
(axis cs:0,0.913238075793812)
--(axis cs:0,0.446761924206188)
--(axis cs:0.1,0.446761924206188)
--(axis cs:0.2,0.490531598666243)
--(axis cs:0.3,0.495500556793564)
--(axis cs:0.4,0.490531598666243)
--(axis cs:0.5,0.427830569293992)
--(axis cs:0.6,0.413831992072823)
--(axis cs:0.7,0.346268439145477)
--(axis cs:0.8,0.341902458362543)
--(axis cs:0.9,0.294969881339626)
--(axis cs:0.9,0.793030118660374)
--(axis cs:0.9,0.793030118660374)
--(axis cs:0.8,0.834097541637457)
--(axis cs:0.7,0.837731560854523)
--(axis cs:0.6,0.890168007927177)
--(axis cs:0.5,0.900169430706008)
--(axis cs:0.4,0.941468401333757)
--(axis cs:0.3,0.944499443206436)
--(axis cs:0.2,0.941468401333757)
--(axis cs:0.1,0.913238075793812)
--(axis cs:0,0.913238075793812)
--cycle;

\addplot [semithick, steelblue31119180]
table {%
0 0.764
0.1 0.712
0.2 0.736
0.3 0.728
0.4 0.7
0.5 0.72
0.6 0.704
0.7 0.668
0.8 0.604
0.9 0.624
};
\addlegendentry{w/ MEE, w/ dist.}

\addplot [semithick, darkorange25512714]
table {%
0 0.68
0.1 0.68
0.2 0.716
0.3 0.72
0.4 0.716
0.5 0.664
0.6 0.652
0.7 0.592
0.8 0.588
0.9 0.544
};
\addlegendentry{w/o MEE, w/ dist.}

\addplot [semithick, steelblue31119180, dashed]
table {%
0 0.772
0.1 0.82
0.2 0.768
0.3 0.832
0.4 0.796
0.5 0.84
0.6 0.764
0.7 0.776
0.8 0.748
0.9 0.676
};
\addlegendentry{w/ MEE, w/o dist.}

\addplot [semithick, darkorange25512714, dashed]
table {%
0 0.752
0.1 0.776
0.2 0.768
0.3 0.78
0.4 0.708
0.5 0.736
0.6 0.712
0.7 0.748
0.8 0.756
0.9 0.64
};
\addlegendentry{w/o MEE, w/o dist.}

\nextgroupplot[
height=5cm,
tick align=inside,
tick pos=left,
width=4cm,
x grid style={darkgray176},
xlabel={},
xmin=0, xmax=2.4,
xtick style={color=black},
y grid style={darkgray176},
ymin=0, ymax=1,
ytick style={color=black},
ytick=\empty,
]
\path [draw=steelblue31119180, fill=steelblue31119180, opacity=0.2]
(axis cs:0,1.05564659966251)
--(axis cs:0,0.784353400337495)
--(axis cs:0.3,0.784353400337495)
--(axis cs:0.6,0.723887539522997)
--(axis cs:0.9,0.61664391486314)
--(axis cs:1.2,0.258032002048262)
--(axis cs:1.5,0.0406828780053869)
--(axis cs:1.8,-0.0471902058765308)
--(axis cs:2.1,-0.0588831137270991)
--(axis cs:2.4,-0.0424426303552648)
--(axis cs:2.4,0.0664426303552648)
--(axis cs:2.4,0.0664426303552648)
--(axis cs:2.1,0.154883113727099)
--(axis cs:1.8,0.263190205876531)
--(axis cs:1.5,0.479317121994613)
--(axis cs:1.2,0.757967997951738)
--(axis cs:0.9,1.00735608513686)
--(axis cs:0.6,1.044112460477)
--(axis cs:0.3,1.05564659966251)
--(axis cs:0,1.05564659966251)
--cycle;

\path [draw=darkorange25512714, fill=darkorange25512714, opacity=0.2]
(axis cs:0,0.86619000805153)
--(axis cs:0,0.38180999194847)
--(axis cs:0.3,0.363901659161722)
--(axis cs:0.6,0.324605578074201)
--(axis cs:0.9,0.316056458039335)
--(axis cs:1.2,0.158651268599163)
--(axis cs:1.5,0.010242861606213)
--(axis cs:1.8,-0.0536474496770062)
--(axis cs:2.1,-0.0585475499463542)
--(axis cs:2.4,-0.0467375485654325)
--(axis cs:2.4,0.0787375485654325)
--(axis cs:2.4,0.0787375485654325)
--(axis cs:2.1,0.146547549946354)
--(axis cs:1.8,0.229647449677006)
--(axis cs:1.5,0.421757138393787)
--(axis cs:1.2,0.649348731400837)
--(axis cs:0.9,0.811943541960665)
--(axis cs:0.6,0.819394421925798)
--(axis cs:0.3,0.852098340838278)
--(axis cs:0,0.86619000805153)
--cycle;

\addplot [semithick, steelblue31119180]
table {%
0 0.92
0.3 0.92
0.6 0.884
0.9 0.812
1.2 0.508
1.5 0.26
1.8 0.108
2.1 0.048
2.4 0.012
};
\addplot [semithick, darkorange25512714]
table {%
0 0.624
0.3 0.608
0.6 0.572
0.9 0.564
1.2 0.404
1.5 0.216
1.8 0.088
2.1 0.044
2.4 0.016
};
\addplot [semithick, steelblue31119180, dashed]
table {%
0 0.864
0.3 0.832
0.6 0.836
0.9 0.764
1.2 0.572
1.5 0.256
1.8 0.072
2.1 0.036
2.4 0.016
};
\addplot [semithick, darkorange25512714, dashed]
table {%
0 0.836
0.3 0.876
0.6 0.824
0.9 0.784
1.2 0.616
1.5 0.332
1.8 0.096
2.1 0.052
2.4 0.016
};
\end{groupplot}

\end{tikzpicture}

%% file: main.bbl
\begin{thebibliography}{10}
\providecommand{\url}[1]{#1}
\csname url@rmstyle\endcsname
\providecommand{\newblock}{\relax}
\providecommand{\bibinfo}[2]{#2}
\providecommand\BIBentrySTDinterwordspacing{\spaceskip=0pt\relax}
\providecommand\BIBentryALTinterwordstretchfactor{4}
\providecommand\BIBentryALTinterwordspacing{\spaceskip=\fontdimen2\font plus
\BIBentryALTinterwordstretchfactor\fontdimen3\font minus \fontdimen4\font\relax}
\providecommand\BIBforeignlanguage[2]{{%
\expandafter\ifx\csname l@#1\endcsname\relax
\typeout{** WARNING: IEEEtran.bst: No hyphenation pattern has been}%
\typeout{** loaded for the language `#1'. Using the pattern for}%
\typeout{** the default language instead.}%
\else
\language=\csname l@#1\endcsname
\fi
#2}}

\bibitem{bhatt2022surprisingly}
A.~Bhatt, A.~Sieler, S.~Puhlmann, and O.~Brock, ``Surprisingly robust in-hand manipulation: An empirical study,'' \emph{arXiv preprint arXiv:2201.11503}, 2022.

\bibitem{kuperberg1990problems}
W.~Kuperberg, ``Problems on polytopes and convex sets,'' in \emph{DIMACS Workshop on polytopes}, 1990, pp. 584--589.

\bibitem{rodriguez2012caging}
A.~Rodriguez, M.~T. Mason, and S.~Ferry, ``From caging to grasping,'' \emph{The International Journal of Robotics Research}, vol.~31, no.~7, pp. 886--900, 2012.

\bibitem{shirizly2024selection}
A.~Shirizly and E.~D. Rimon, ``Selection of secure gravity based caging grasps of planar objects: Robustness and experimental validation,'' \emph{IEEE Transactions on Robotics}, 2024.

\bibitem{varava2017herding}
A.~Varava, K.~Hang, D.~Kragic, and F.~T. Pokorny, ``Herding by caging: a topological approach towards guiding moving agents via mobile robots.'' in \emph{Robotics: Science and Systems}, 2017, pp. 1--9.

\bibitem{bircher2021complex}
W.~G. Bircher, A.~S. Morgan, and A.~M. Dollar, ``Complex manipulation with a simple robotic hand through contact breaking and caging,'' \emph{Science Robotics}, vol.~6, no.~54, p. eabd2666, 2021.

\bibitem{beddow2021caging}
L.~Beddow, H.~Wurdemann, and D.~Kanoulas, ``A caging inspired gripper using flexible fingers and a movable palm,'' in \emph{2021 IEEE/RSJ International Conference on Intelligent Robots and Systems (IROS)}.\hskip 1em plus 0.5em minus 0.4em\relax IEEE, 2021, pp. 7195--7200.

\bibitem{bircher2019design}
W.~G. Bircher and A.~M. Dollar, ``Design principles and optimization of a planar underactuated hand for caging grasps,'' in \emph{2019 International Conference on Robotics and Automation (ICRA)}.\hskip 1em plus 0.5em minus 0.4em\relax IEEE, 2019, pp. 1608--1613.

\bibitem{wang2024caging}
G.~Wang, K.~Ren, A.~S. Morgan, and K.~Hang, ``Caging in time: A framework for robust object manipulation under uncertainties and limited robot perception,'' \emph{arXiv preprint arXiv:2410.16481}, 2024.

\bibitem{mahler2018synthesis}
J.~Mahler, F.~T. Pokorny, S.~Niyaz, and K.~Goldberg, ``Synthesis of energy-bounded planar caging grasps using persistent homology,'' \emph{IEEE Transactions on Automation Science and Engineering}, vol.~15, no.~3, pp. 908--918, 2018.

\bibitem{dong2024quasi}
Y.~Dong and F.~T. Pokorny, ``Quasi-static soft fixture analysis of rigid and deformable objects,'' in \emph{Proc. Int. Conf. Robot. Automat.}\hskip 1em plus 0.5em minus 0.4em\relax IEEE, 2024, pp. 6513--6520.

\bibitem{ferrari1992planning}
C.~Ferrari, J.~F. Canny, \emph{et~al.}, ``Planning optimal grasps.'' in \emph{Proc. Int. Conf. Robot. Automat.}, vol.~3, no.~4, 1992, p.~6.

\bibitem{pollard1996synthesizing}
N.~S. Pollard, ``Synthesizing grasps from generalized prototypes,'' in \emph{Proc. Int. Conf. Robot. Automat.}, vol.~3.\hskip 1em plus 0.5em minus 0.4em\relax IEEE, 1996, pp. 2124--2130.

\bibitem{roa2015grasp}
M.~A. Roa and R.~Su{\'a}rez, ``Grasp quality measures: review and performance,'' \emph{Autonomous robots}, vol.~38, pp. 65--88, 2015.

\bibitem{saxena2008learning}
A.~Saxena, L.~L. Wong, and A.~Y. Ng, ``Learning grasp strategies with partial shape information.'' in \emph{AAAI}, vol.~3, no.~2, 2008, pp. 1491--1494.

\bibitem{mahler2017dex}
J.~Mahler, J.~Liang, S.~Niyaz, M.~Laskey, R.~Doan, X.~Liu, J.~A. Ojea, and K.~Goldberg, ``Dex-net 2.0: Deep learning to plan robust grasps with synthetic point clouds and analytic grasp metrics,'' \emph{arXiv preprint arXiv:1703.09312}, 2017.

\bibitem{mahler2018dex}
J.~Mahler, M.~Matl, X.~Liu, A.~Li, D.~Gealy, and K.~Goldberg, ``Dex-net 3.0: Computing robust vacuum suction grasp targets in point clouds using a new analytic model and deep learning,'' in \emph{2018 IEEE International Conference on robotics and automation (ICRA)}.\hskip 1em plus 0.5em minus 0.4em\relax IEEE, 2018, pp. 5620--5627.

\bibitem{wang2020feature}
C.~Wang, X.~Zhang, X.~Zang, Y.~Liu, G.~Ding, W.~Yin, and J.~Zhao, ``Feature sensing and robotic grasping of objects with uncertain information: A review,'' \emph{Sensors}, vol.~20, no.~13, p. 3707, 2020.

\bibitem{bohg2013data}
J.~Bohg, A.~Morales, T.~Asfour, and D.~Kragic, ``Data-driven grasp synthesis—a survey,'' \emph{IEEE Transactions on robotics}, vol.~30, no.~2, pp. 289--309, 2013.

\bibitem{daniels2023grasping}
A.~Daniels, S.~Kerz, S.~Bari, V.~Gabler, and D.~Wollherr, ``Grasping in uncertain environments: A case study for industrial robotic recycling,'' in \emph{2023 IEEE International Conference on Systems, Man, and Cybernetics (SMC)}.\hskip 1em plus 0.5em minus 0.4em\relax IEEE, 2023, pp. 3514--3521.

\bibitem{andrychowicz2020learning}
O.~M. Andrychowicz, B.~Baker, M.~Chociej, R.~Jozefowicz, B.~McGrew, J.~Pachocki, A.~Petron, M.~Plappert, G.~Powell, A.~Ray, \emph{et~al.}, ``Learning dexterous in-hand manipulation,'' \emph{The International Journal of Robotics Research}, vol.~39, no.~1, pp. 3--20, 2020.

\bibitem{jankowski2024planning}
J.~Jankowski, L.~Bruderm{\"u}ller, N.~Hawes, and S.~Calinon, ``Planning for robust open-loop pushing: Exploiting quasi-static belief dynamics and contact-informed optimization,'' \emph{arXiv preprint arXiv:2404.02795}, 2024.

\bibitem{rimon1999caging}
E.~Rimon and A.~Blake, ``Caging planar bodies by one-parameter two-fingered gripping systems,'' \emph{The International Journal of Robotics Research}, vol.~18, no.~3, pp. 299--318, 1999.

\bibitem{dong2024characterizing}
Y.~Dong, X.~Cheng, and F.~T. Pokorny, ``Characterizing manipulation robustness through energy margin and caging analysis,'' \emph{IEEE Robot. Automat. Lett.}, 2024.

\bibitem{zhang2024introduction}
Y.~Zhang, P.~Khanduri, I.~Tsaknakis, Y.~Yao, M.~Hong, and S.~Liu, ``An introduction to bilevel optimization: Foundations and applications in signal processing and machine learning,'' \emph{IEEE Signal Processing Magazine}, vol.~41, no.~1, pp. 38--59, 2024.

\bibitem{wang2024diffusebot}
T.-H.~J. Wang, J.~Zheng, P.~Ma, Y.~Du, B.~Kim, A.~Spielberg, J.~Tenenbaum, C.~Gan, and D.~Rus, ``Diffusebot: Breeding soft robots with physics-augmented generative diffusion models,'' \emph{Advances in Neural Information Processing Systems}, vol.~36, 2024.

\bibitem{bhatia2021evolution}
J.~Bhatia, H.~Jackson, Y.~Tian, J.~Xu, and W.~Matusik, ``Evolution gym: A large-scale benchmark for evolving soft robots,'' \emph{Advances in Neural Information Processing Systems}, vol.~34, pp. 2201--2214, 2021.

\bibitem{kim2021mo}
Y.~Kim, Z.~Pan, and K.~Hauser, ``Mo-bbo: Multi-objective bilevel bayesian optimization for robot and behavior co-design,'' in \emph{2021 IEEE International Conference on Robotics and Automation (ICRA)}.\hskip 1em plus 0.5em minus 0.4em\relax IEEE, 2021, pp. 9877--9883.

\bibitem{liu2023learning}
Z.~Liu, S.~Tian, M.~Guo, K.~Liu, and J.~Wu, ``Learning to design and use tools for robotic manipulation,'' in \emph{7th Annual Conference on Robot Learning}, 2023.

\bibitem{schneider2024task}
R.~Schneider, D.~Honerkamp, T.~Welschehold, and A.~Valada, ``Task-driven co-design of mobile manipulators,'' \emph{arXiv preprint arXiv:2412.16635}, 2024.

\bibitem{rajani2023co}
C.~Rajani, K.~Arndt, D.~Blanco-Mulero, K.~S. Luck, and V.~Kyrki, ``Co-imitation: learning design and behaviour by imitation,'' in \emph{Proceedings of the AAAI Conference on Artificial Intelligence}, vol.~37, no.~5, 2023, pp. 6200--6208.

\bibitem{yuan2022transform2act}
Y.~Yuan, Y.~Song, Z.~Luo, W.~Sun, and K.~M. Kitani, ``Transform2act: Learning a transform-and-control policy for efficient agent design,'' in \emph{International Conference on Learning Representations}, 2022.

\bibitem{jackson2021orchid}
L.~Jackson, C.~Walters, S.~Eckersley, P.~Senior, and S.~Hadfield, ``Orchid: optimisation of robotic control and hardware in design using reinforcement learning,'' in \emph{2021 IEEE/RSJ International Conference on Intelligent Robots and Systems (IROS)}.\hskip 1em plus 0.5em minus 0.4em\relax IEEE, 2021, pp. 4911--4917.

\bibitem{liu2024paperbot}
R.~Liu, J.~Liang, S.~Sudhakar, H.~Ha, C.~Chi, S.~Song, and C.~Vondrick, ``Paperbot: Learning to design real-world tools using paper,'' \emph{arXiv preprint arXiv:2403.09566}, 2024.

\bibitem{guo2024learning}
M.~Guo, Z.~Liu, S.~Tian, Z.~Xie, J.~Wu, and C.~K. Liu, ``Learning to design 3d printable adaptations on everyday objects for robot manipulation,'' in \emph{2024 IEEE International Conference on Robotics and Automation (ICRA)}.\hskip 1em plus 0.5em minus 0.4em\relax IEEE, 2024, pp. 824--830.

\bibitem{pan2021emergent}
X.~Pan, A.~Garg, A.~Anandkumar, and Y.~Zhu, ``Emergent hand morphology and control from optimizing robust grasps of diverse objects,'' in \emph{2021 IEEE International Conference on Robotics and Automation (ICRA)}.\hskip 1em plus 0.5em minus 0.4em\relax IEEE, 2021, pp. 7540--7547.

\bibitem{xu2024dynamics}
X.~Xu, H.~Ha, and S.~Song, ``Dynamics-guided diffusion model for robot manipulator design,'' \emph{arXiv preprint arXiv:2402.15038}, 2024.

\bibitem{swersky2013multi}
K.~Swersky, J.~Snoek, and R.~P. Adams, ``Multi-task bayesian optimization,'' \emph{Advances in neural information processing systems}, vol.~26, 2013.

\bibitem{makapunyo2012measurement}
T.~Makapunyo, T.~Phoka, P.~Pipattanasomporn, N.~Niparnan, and A.~Sudsang, ``Measurement framework of partial cage quality,'' in \emph{2012 IEEE international conference on robotics and biomimetics (ROBIO)}.\hskip 1em plus 0.5em minus 0.4em\relax IEEE, 2012, pp. 1812--1816.

\bibitem{gammell2015batch}
J.~D. Gammell, S.~S. Srinivasa, and T.~D. Barfoot, ``Batch informed trees (bit*): Sampling-based optimal planning via the heuristically guided search of implicit random geometric graphs,'' in \emph{2015 IEEE international conference on robotics and automation (ICRA)}.\hskip 1em plus 0.5em minus 0.4em\relax IEEE, 2015, pp. 3067--3074.

\bibitem{shirai2024robust}
Y.~Shirai, D.~K. Jha, and A.~U. Raghunathan, ``Robust pivoting manipulation using contact implicit bilevel optimization,'' \emph{IEEE Transactions on Robotics}, 2024.

\bibitem{schulman2017proximal}
J.~Schulman, F.~Wolski, P.~Dhariwal, A.~Radford, and O.~Klimov, ``Proximal policy optimization algorithms,'' \emph{arXiv preprint arXiv:1707.06347}, 2017.

\bibitem{bonilla2007multi}
E.~V. Bonilla, K.~Chai, and C.~Williams, ``Multi-task gaussian process prediction,'' \emph{Advances in neural information processing systems}, vol.~20, 2007.

\bibitem{garivier2011upper}
A.~Garivier and E.~Moulines, ``On upper-confidence bound policies for switching bandit problems,'' in \emph{International conference on algorithmic learning theory}.\hskip 1em plus 0.5em minus 0.4em\relax Springer, 2011, pp. 174--188.

\bibitem{coumans2016pybullet}
E.~Coumans and Y.~Bai, ``Pybullet, a python module for physics simulation for games, robotics and machine learning,'' 2016.

\bibitem{catto_box2d}
E.~Catto, ``Box2d,'' \url{https://box2d.org/}.

\bibitem{kramer2017genetic}
O.~Kramer and O.~Kramer, \emph{Genetic algorithms}.\hskip 1em plus 0.5em minus 0.4em\relax Springer, 2017.

\bibitem{srinivas2009gaussian}
N.~Srinivas, A.~Krause, S.~M. Kakade, and M.~Seeger, ``Gaussian process optimization in the bandit setting: No regret and experimental design,'' \emph{arXiv preprint arXiv:0912.3995}, 2009.

\end{thebibliography}
